\documentclass[twoside]{article}

\usepackage[preprint]{aistats2020}
\usepackage[round]{natbib}

\usepackage[symbol]{footmisc}

\usepackage[T1]{fontenc}    
\usepackage{hyperref}       
\usepackage{url}            
\usepackage{booktabs}       
\usepackage{amsfonts,amsmath,amssymb,amsthm}       
\usepackage{nicefrac}       
\usepackage{siunitx}
\usepackage{graphicx}
\usepackage{listings}
\usepackage{dsfont}
\usepackage{bm}

\usepackage{color}
\usepackage{wrapfig}
\usepackage[ruled]{algorithm2e}

\newcommand\cut[1]{}

\newcommand{\squishlist}{
   \begin{list}{$\bullet$}
    { \setlength{\itemsep}{0pt}      \setlength{\parsep}{3pt}
      \setlength{\topsep}{3pt}       \setlength{\partopsep}{0pt}
      \setlength{\leftmargin}{1.5em} \setlength{\labelwidth}{1em}
      \setlength{\labelsep}{0.5em} } }

\newcommand{\squishlisttwo}{
   \begin{list}{$\bullet$}
    { \setlength{\itemsep}{0pt}    \setlength{\parsep}{0pt}
      \setlength{\topsep}{0pt}     \setlength{\partopsep}{0pt}
      \setlength{\leftmargin}{2em} \setlength{\labelwidth}{1.5em}
      \setlength{\labelsep}{0.5em} } }

\newcommand{\squishend}{
    \end{list}  }
 
{}
{}
{}
{}
{}
{}

\newcommand{\myvec}[1]{\mbox{$\mathbf{#1}$}}

\newcommand{\vx}{\mbox{$\myvec{x}$}}

\newcommand{\vy}{\mbox{$\myvec{y}$}}

\newcommand{\be}{\begin{equation}}
\newcommand{\ee}{\end{equation}}
\newcommand{\bea}{\begin{eqnarray}}
\newcommand{\eea}{\end{eqnarray}}
\newcommand{\beaa}{\begin{eqnarray*}}
\newcommand{\eeaa}{\end{eqnarray*}}

\DeclareFixedFont{\ttb}{T1}{txtt}{bx}{n}{8} 
\DeclareFixedFont{\ttm}{T1}{txtt}{m}{n}{8}  
\definecolor{deepblue}{rgb}{0,0,0.5}
\definecolor{deepred}{rgb}{0.6,0,0}
\definecolor{deepgreen}{rgb}{0,0.5,0}

\usepackage[export]{adjustbox}
\usepackage{subfig}

\begin{document}
\lstset{
language=Python,
backgroundcolor=\color{white},
basicstyle=\ttm,
keywordstyle=\ttb\color{deepgreen},
emph={MyClass,__init__},          
emphstyle=\ttb\color{deepred},    
stringstyle=\color{deepblue},
commentstyle=\color{red},
frame=tb,
showstringspaces=false,
literate=
{sample}{{{\color{deepblue}sample{}}}}5
{observe}{{{\color{deepblue}observe{}}}}6
}
%

%

\twocolumn[

\aistatstitle{Attention for Inference Compilation}

\aistatsauthor{
  William Harvey\footnotemark[2]
  \And
  Andreas Munk\footnotemark[2]
  \And
  At{\i}l{\i}m G{\"u}ne{\c s} Baydin}

\aistatsaddress{ 
  Dept. of Computer Science\\
  University of British Columbia\\
  Vancouver, B.C.\\
  Canada \\
  \texttt{wsgh@cs.ubc.ca}
  \And
  Dept. of Computer Science\\
  University of British Columbia\\
  Vancouver, B.C.\\
  Canada \\
  \texttt{amunk@cs.ubc.ca}
  \And
  Dept. of Engineering Science\\
  University of Oxford\\
  Oxford, OX2 6PN \\
  United Kingdom\\
  \texttt{gunes@robots.ox.ac.uk}
}

\aistatsauthor{
  Alexander Bergholm
  \And
  Frank Wood }

\aistatsaddress{
  Dept. of Computer Science\\
  University of British Columbia\\
  Vancouver, B.C.\\
  Canada \\
  \texttt{bergholm.alexander@gmail.com}
  \And 
  Dept. of Computer Science\\
  University of British Columbia\\
  Vancouver, B.C.\\
  Canada \\
  \texttt{fwood@cs.ubc.ca}
}
]

\begin{abstract}
  We present a new approach to automatic amortized inference in universal
  probabilistic programs which improves performance compared to current methods.
  Our approach is a variation of inference compilation (IC) which leverages deep
  neural networks to approximate a posterior distribution over latent variables
  in a probabilistic program. A challenge with existing IC network architectures
  is that they can fail to model long-range dependencies between latent
  variables. To address this, we introduce an attention mechanism that attends
  to the most salient variables previously sampled in the execution of a
  probabilistic program. We demonstrate that the addition of attention allows
  the proposal distributions to better match the true posterior, enhancing
  inference about latent variables in simulators.
\end{abstract}

\section{INTRODUCTION}
Probabilistic programming
languages~\citep{van2018introduction,mansinghka2014venture,milch2005a,wood2014new,pfeffer2009figaro,InferNET18,goodman2008a,gordon2014a}
allow for automatic inference about random variables in generative models
written as programs. Conditions on these random variables are imposed through
\texttt{observe} statements, while the \texttt{sample} statements define latent
variables we seek to draw inference about. Common to the different languages is
the existence of an inference backend, which contains one or more general
inference methods.

Recent research has addressed the task of making repeated inference less
computationally expensive, by using up-front computation to reduce the cost
of later executions, an approach known as amortized
inference~\citep{gershman2014amortized}. One new method called inference
compilation (IC) \citep{le2016inference} enables fast inference on arbitrarily
complex and non-differentiable generative models. The approximate posterior
distribution it learns can be combined with importance sampling at inference
time, so that inference is asymptotically correct. It has been successfully used
for Captcha solving~\citep{le2016inference} and inference in particle physics
simulators~\citep{baydin2018efficient}.

The neural network used in IC is trained to approximate the joint posterior
given the observed variables by sequentially proposing a distribution for each
latent variable generated during an execution of a program. As such, capturing
the possible dependencies on previously sampled variables is essential to
achieve good performance. IC uses a Long Short Term Memory (LSTM)-based
architecture~\citep{hochreiter1997a} to encapsulate these dependencies. However, this architecture fails to learn the
dependency between highly dependent random variables when they are sampled far
apart (with several other variables sampled in-between). This motivates allowing
the neural network which parameterizes the proposal distribution for each latent
variable to explicitly access any previously sampled variables. Inspired by the
promising results of attention for tasks involving long-range
dependencies~\citep{jaderberg2015spatial, vaswani2017attention,
  seo2016bidirectional}, we implemented an attention mechanism over previously
sampled values. This enables the network to selectively attend to any
combination of previously sampled values, regardless of their order and the
trace length. We show that our approach significantly improves the approximation
of the posterior, and hence facilitates faster inference.

The principle contributions of this paper are two-fold: we show that attention
improves the performance of IC, and we show that we are able to use IC to
perform fault detection via amortized inference in complex simulators.
Section~\ref{sec:background} introduces the concepts of probabilistic
programming, inference compilation and attention for neural networks. We then
describe our approach in Section~\ref{sec:method} and our experiments in
Section~\ref{sec:experiments}.

\section{BACKGROUND}
\label{sec:background}
\begin{figure}[t]
  \centering
  \includegraphics[width=0.15\textwidth]{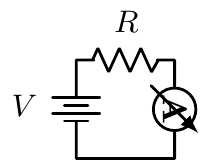}
  \caption{The electric circuit modelled by the probabilistic program
    in Figure~\ref{fig:simple_program}.}
  \label{fig:simple_circuit}
\end{figure}

\subsection{Probabilistic Programming}
Probabilistic programming languages (PPLs) allow the specification of
probabilistic generative models (and therefore probability distributions) as
computer programs. Universal PPLs, which are based on Turing
complete languages, may express models with an unbounded number of
random variables. To this end, they combine traditional programming languages
with the ability to sample a latent random variable (using syntax which we
denote as a \texttt{sample} statement) and to condition these latent
variables on the values of other, observed, random variables (using an
\texttt{observe} statement).
More formally, following~\citep{le2016inference}, we will operate on higher-order probabilistic programs in which we discuss the joint distribution of variables in an execution ``trace'' $(x_t,
a_t, i_t)$, where $t=1,\dots, T$, with $T$ being the trace length (which may
vary between executions). $x_t$ denotes the value sampled at the $t$th
\texttt{sample} statement encountered, $a_t$ is the address of this
\texttt{sample} statement and $i_t$ represents the instance: the number of times
the same address has been encountered previously, i.e.
$i_t=\sum_{j=i}^t\mathds{1}(a_t=a_j)$. We shall assume that there is a fixed
number of observations, $N$, and these are denoted by $\vy=(y_1,\dots,y_N)$, and
we denote the latent variables as $\vx = (x_1,\dots,x_T)$. Using this formalism,
we express the joint distribution of a trace and observations as,
\begin{equation}\label{eq:pp-joint}
    p(\vx,\vy)=\prod_{t=1}^T f_{a_t}(x_t|x_{1:t-1})\prod_{n=1}^N g_n(y_n|x_{1:\tau(n)}),
\end{equation}
where $f_{a_t}$ is the probability distribution specified by the \texttt{sample}
statement at address $a_t$, and $g_n$ is the probability distribution specified
by the $n$th \texttt{observe} statement. A mapping from the index, $n$, of the
\texttt{observe} statement to the index of the most recent \texttt{sample}
statement before the $n$th \texttt{observe} statement, is denoted by $\tau$.

As an example, consider the simple circuit as well as the probabilistic program shown in Figure~\ref{fig:simple_circuit}, which expresses the joint distribution over the battery voltage, $V$, whether the resistor is faulty, $F$, the resistance of the
resistor, $R$, and the measured current, $I$, as $p(V,F,R,I) =
p(I|V,R)p(R|F)p(F)p(V)$.

\begin{figure}[t]
  \centering
  \includegraphics[width=0.45\textwidth]{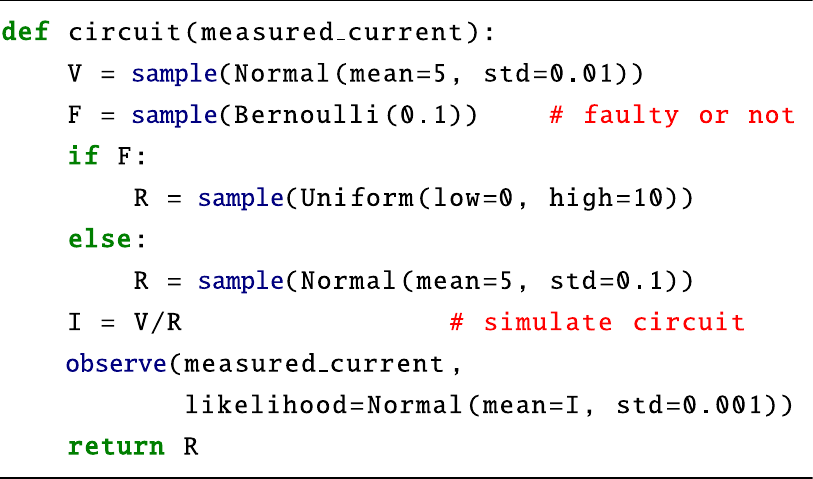}
  \caption{Probabilistic program modeling the circuit
    in Figure~\ref{fig:simple_circuit} with a possibly faulty resistor. First the
    voltage, $V$, of the battery is sampled from a Gaussian prior centered on
    $5\si{\volt}$. We then sample whether or not the resistor is faulty.
    If it is, its value is sampled from a broad uniform distribution. Otherwise, its value is sampled from a tightly peaked Gaussian. A noisy measurement of the current is then sampled from a Gaussian prior centered on the true value.}%
  \label{fig:simple_program}%
\end{figure}

Traces will have the form $(x_t,a_t,i_t)_{t=1}^{T=3}$ where there are two trace
``types,'' one corresponding to the sequence of addresses of random variables
generated if the resistor is faulty, and the other the opposite. In other words
$a_1$ is the address where $V$ is sampled, $a_2$ is the address where $F$ is
sampled, and $a_3$ is the address from which $R$ is sampled, which depends on
$F$. The instance counts in this program are always $i_1=i_2=i_3=1$, and the
observation, \texttt{measured\_current} $\sim \mathcal{N}(I, 0.001)$, with
$N=1$.

This generative model allows posterior inference to be performed over the joint
distribution of the input voltage $V$, current $I$, and ``faulty'' variable $F$
given the observed \texttt{measured\_current}. Estimates of the marginal
posterior distribution over $F$ make it possible to directly answer questions
such as whether the resistor is faulty or not. We will return to a more complex
version of this problem in section~\ref{sec:circuits}. Generally, PPLs are
designed to infer posterior distributions over the latent variables given the
observations. Inference in probabilistic programs is carried out with algorithms
such as Sequential Importance Sampling (SIS)~\citep{arulampalam2002a},
Lightweight Metropolis-Hastings~\citep{wingate2011a}, and Sequential Monte
Carlo~\citep{del2006sequential}. However, these algorithms are too
computationally expensive for use in real-time applications. Therefore, recent
research~\citep{le2016inference, kulkarni2015picture} has considered amortizing
the computational cost by performing up-front computation (for a given model) to
allow faster inference later (given this model and any observed values).

\begin{figure*}[t!]
  \vskip 0.2in
    \centering
    \includegraphics[width=0.75\textwidth]{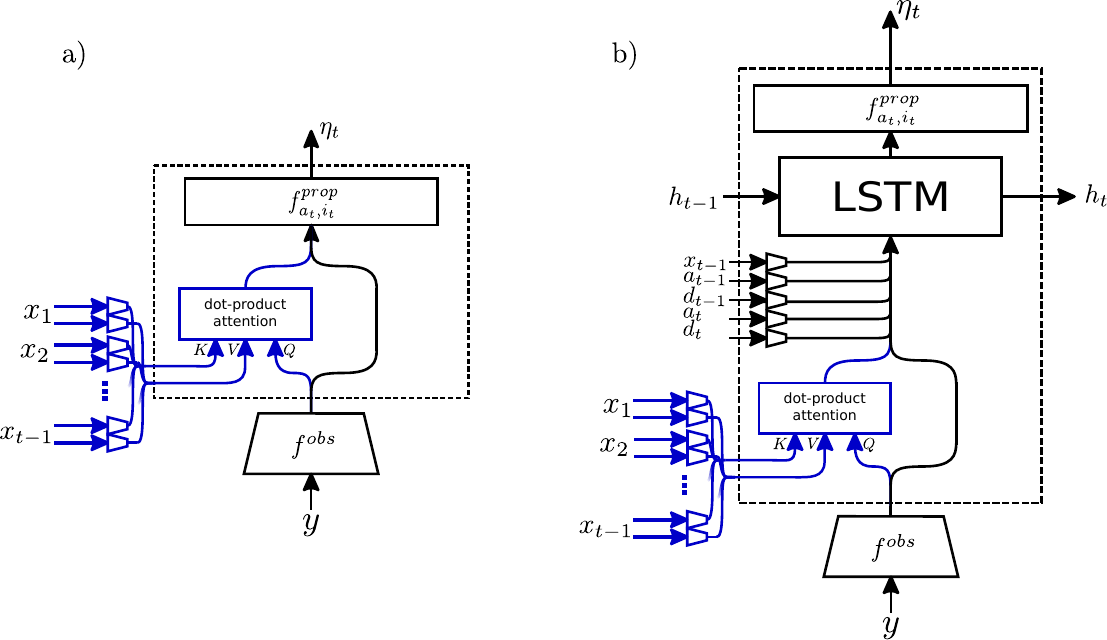}
    \caption{Feedforward and LSTM neural network architectures with attention
      mechanisms. The components inside the dashed line are run once at each
      \texttt{sample} statement in a program trace, while the parts outside this
      line are only run once per trace. The attention mechanism is denoted in
      blue.}
    \label{fig:architectures}
  \vskip -0.2in
\end{figure*}

\subsection{Inference Compilation}

IC~\citep{le2016inference} is a generalized method for
performing amortized inference in the framework of universal probabilistic programming.
It involves training neural networks, which we describe as ``inference networks,'' whose outputs parameterize proposal distributions used for SIS.

\subsubsection{Objective Function}

In IC, we desire to match the proposal distribution, $q(\vx|\vy;\phi) =
\prod_{t=1}^T q_{a_t, i_t}(x_t | \eta_t(x_{1:t-1}, \vy, \phi))$, closely to the true posterior, $p(\vx|\vy)$. The Kullback-Leibler divergence,
$D_\text{KL}(p(\vx|\vy)||q(\vx|\vy;\phi))$, is used as a measure of this
``closeness''. In order to ensure over any observed $\vy$, an expectation of this divergence is taken with respect to $p(\vy)$,
\begin{align}
&\begin{aligned}
\mathcal{L}(\phi)&=\mathbb{E}_{p(\vy)}[D_\text{KL}(p(\vx|\vy)||q(\vx|\vy;\phi))] \\
      & =\mathbb{E}_{p(\vx,\vy)}[-\log q(\vx|\vy,\phi)] + \text{const}
      \end{aligned}
\label{eq:loss}
\end{align}
where the joint distribution takes the form of Eq.~\eqref{eq:pp-joint}.

The parameters, $\phi$, are updated using
gradient descent using the following
estimate of the gradient of \eqref{eq:loss},
\begin{equation}\label{eq:est_loss}
   \setlength\abovedisplayskip{0pt}
   \setlength\belowdisplayskip{0pt}
    \nabla_\phi\mathcal{L}(\phi)\approx\frac{1}{M}\sum_{m=1}^M-\nabla_\phi\log q(\vx^m|\vy^m,\phi),
\end{equation}
where $(\vx^m, \vy^m)\sim p(\vx,\vy)$ for $m=\{1,\dots,M\}$. Note that the loss used, and the estimates of the gradients, are identical to those in the sleep-phase of wake-sleep~\citep{hinton1995wake}.

\subsubsection{Architecture}
\label{sec:std-IC}

The architecture used in IC~\citep{baydin2018efficient,le2016inference} consists of the black components shown in
Figure~\ref{fig:architectures}b. Before performing inference, observations $\vy$
are embedded by a learned \texttt{observe} embedder, $f^{obs}$. At each
\texttt{sample} statement encountered as the program is run, the LSTM is run for
one time step. It receives an input consisting of the concatenation of the embedding of the observed values, $f^{obs}(\vy)$, an embedding of $x_{t-1}$, the value sampled at the previous \texttt{sample} statement, embeddings of the current address, instance and distribution-type, denoted $a_t$, $i_t$ and $d_t$ respectively, embeddings of the previous address, instance and distribution type: $a_{t-1}$, $i_{t-1}$ and $d_{t-1}$. 

The embedder used for $x_{t-1}$ is specific to $(a_{t-1}, i_{t-1})$, the address
and instance from which $x_{t-1}$ was sampled. The output of the LSTM is fed
into a proposal layer, which is specific to the address and instance ($a_t$ and
$i_t$). The proposal layer outputs the parameters, $\eta_t$, of a proposal
distribution for the variable at this \texttt{sample} statement.
\subsubsection{Inference using Sequential Importance Sampling}

In IC inference is performed by SIS~\citep{arulampalam2002a,doucet2009tutorial}, which is compatible with
latent variable inference in higher-order probabilistic programs~\citep{wood2014new}. SIS
produces a set of $K$ weighted samples $\{(\vx_k,w_k)\}_{k=1}^K$, such that the
posterior and expectations of a function $g(\vx)$ are approximated by
\begin{align}
&\begin{aligned}
    p(\vx|\vy)&\approx\hat{p}(\vx|\vy)=\frac{\sum_{k=1}^Kw_k\delta(\vx_k-\vx)}{\sum_{k=1}^Kw_k} \\
    &\mathbb{E}[g(\vx)]\approx\frac{\sum_{k=1}^Kw_kg(\vx_k)}{\sum_{k=1}^Kw_k},
    \end{aligned}
\end{align}
where $\delta$ is the Dirac delta function. Each weight $w_k$ is calculated for
each trace $\vx_k$ according to
\begin{equation*}
    w_k=\prod_{n=1}^Ng_n(y_n|x^k_{1:\tau_{k}(n)})\prod_{t=1}^{T_k}\frac{f_{a_t}(x_t^k|x_{1:t-1}^k)}{q_{a_t,i_t}(x_t^k|x_{1:t-1}^k)},
\end{equation*}
where $T_k$ denotes the $k$th trace length with $k=\{1,\dots,K\}$, and
$q_{a_t,i_t}$ is given by the inference network learned by minimizing the loss
function~\eqref{eq:est_loss}.

\subsection{Dot-product Attention}
\label{sec:dpa}

Attention has recently been shown to be useful in a number of tasks,
including image captioning, machine translation, and image
generation~\citep{xu2015show, bahdanau2014neural, gregor2015draw}. The two broad
types of attention are hard and soft attention. Hard
attention~\citep{ba2014multiple, xu2015show} selects a single ``location'' to
attend to, and thus requires only this location to be embedded. However, it is
non-differentiable. In contrast, soft attention
mechanisms~\citep{vaswani2017attention, xu2015show} are fully differentiable.
They typically require splitting the input into a finite number of locations. Each location is embedded separately, and a
weighted average of these embeddings is returned as the output.

Our proposed architecture incorporates \textit{dot-product}
attention~\citep{luong2015effective, vaswani2017attention}, a form of soft
attention. This choice is justified by the embeddings of the value sampled at
time step $t$ being used at all later time steps $t+1,\dots,T$. This means that
the computational cost of calculating the embeddings scales linearly with the
trace length. Since this is no worse than the rate that hard attention achieves,
we select soft attention for its differentiability. The specific use of
dot-product attention is due to the efficiency of the calculation of attention
weights. Each trace requires the computation of $\mathcal{O}(T^2)$ attention
weights and some programs may contain thousands of \texttt{sample}
statements~\citep{baydin2018efficient} so fast weight computation is paramount.

\begin{figure}[t]
  \vskip 0.1in
  \centering
  \includegraphics[width=0.4\textwidth]{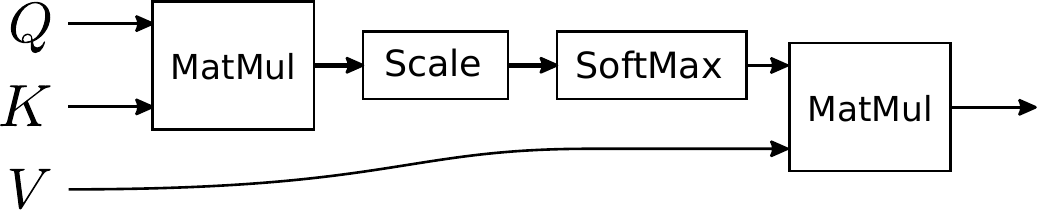}
  \caption{A scaled dot-product attention mechanism. Figure
    adapted from \citep{vaswani2017attention}}
  \label{fig:dpa}
\end{figure}
The dot-product attention module \citep{luong2015effective,vaswani2017attention},
shown in Figure~\ref{fig:dpa}, receives three inputs: one or more query vectors
(which describe the desired properties of the locations to attend to), a key
vector for each location, and a value vector for each location. In Figure 3 in
the Appendix, these are represented as the matrices $Q \in \mathbb{R}^{q \times
  k}$, $K \in \mathbb{R}^{k \times l}$, and $V \in \mathbb{R}^{l \times v}$,
respectively. Here, $l$ is the number of locations, $k$ is the length of each
query and key embedding, $v$ is the length of each value embedding, and $q$ is
the number of queries. For each query, attention weights are computed for every
location by taking the dot-product of the query vector and the relevant key. A
SoftMax is then applied to ensure that the sum of the weights over every
location is 1 (for each query). These weights are used to compute weighted
averages of the values. The output of the attention mechanism is a concatenation
of these. This procedure can be performed efficiently using matrix
multiplications, and so calculating the attention weights is more
computationally efficient than in other types of soft attention (which typically
differ in how the weights are calculated)~\citep{bahdanau2014neural}. In scaled
dot-product attention~\citep{vaswani2017attention}, the product $QK$ is
multiplied by a scalar before performing the softmax. This scalar is chosen to
ensure that the output of the softmax is not saturated upon initialising the
weights, so the gradients propagated through are large enough for effective
training.

\section{METHOD}
\label{sec:method}

\begin{figure*}[t!]
   \vskip 0.2in \centering
   \includegraphics[scale=0.65]{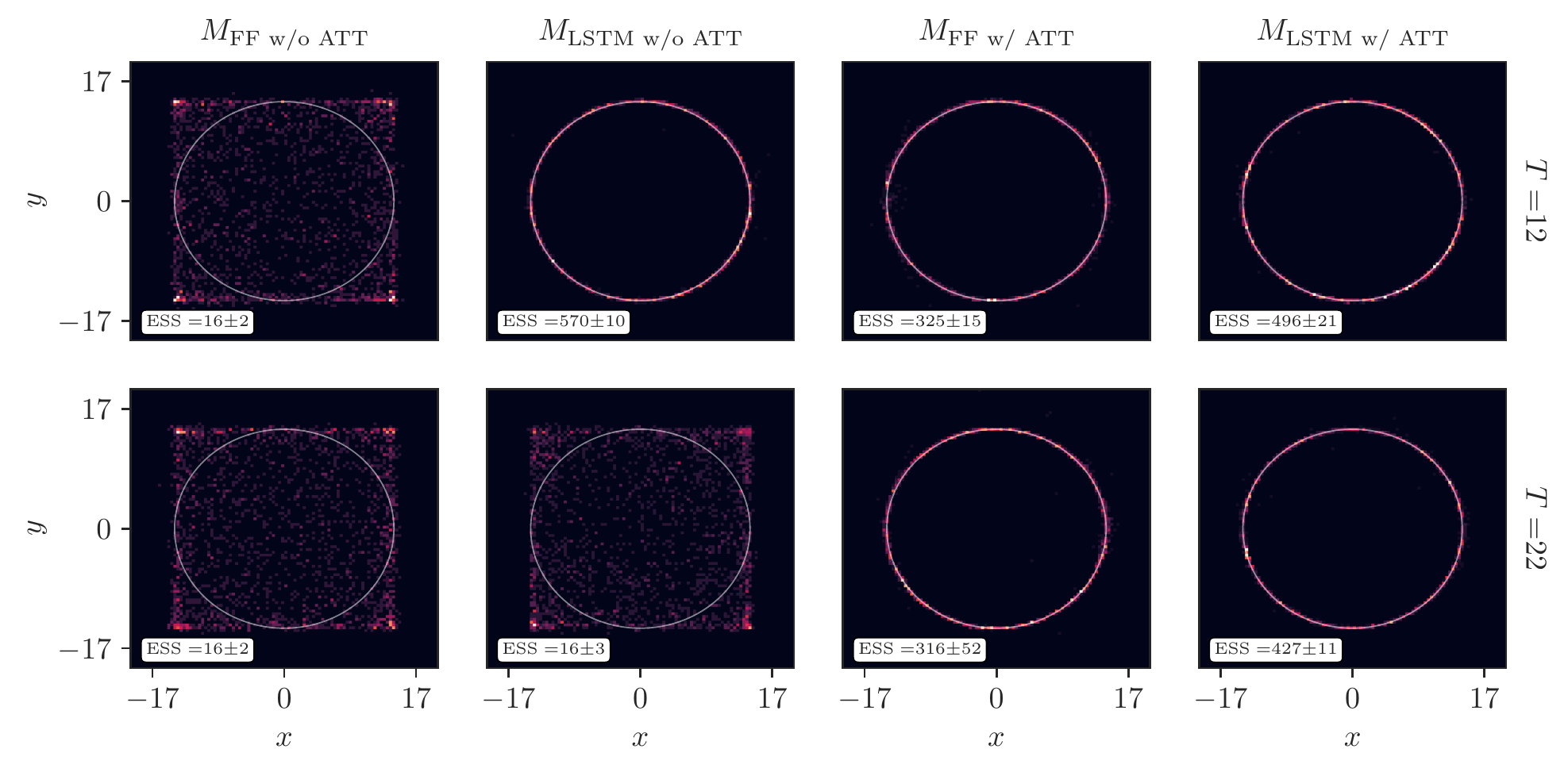}
   \caption{Different IC neural network architectures, with
     and without attention, showing improved IC performance
     for those with attention on an inference task involving highly correlated
     latent random variables. Each image shows 2000 posterior samples of $x$ and $y$ from
     the model described in section 4 conditioned on $\hat{r}^2 = 200$. The top row contains samples from program with 10 nuisance
     variables, i.e. trace length $T=12$. The bottom row contains samples
     from the model with 20 nuisance variables, i.e. trace length
     $T=22$. The white circles indicate the mode of the true posterior. $M_\text{FF w/o ATT}$ is unable to learn any dependency
     between $x$ and $y$. With 10 nuisance variables, we see that attention
     provides no advantage over the standard LSTM IC architecture. However, as the number
     of nuisance variables increases from 10 to 20, attention becomes beneficial,
     maintaining a high effective sample size (ESS) while the performance of
     $M_\text{LSTM w/o ATT}$ reduces to that of $M_\text{FF w/o ATT}$. The
     results were similar when different radii were observed. The ESS in the figure is calculated by averaging 10 estimates.}
  \vskip -0.2in
    \label{fig:circ_dist}
\end{figure*} 

We compare against two ``baseline architectures'' in all experiments: a
feedforward network learned for each address and instance pair $(a, i)$; and the
LSTM-based architecture described in section~\ref{sec:std-IC}. These are shown
as the black components of Figure~\ref{fig:architectures}a and
\ref{fig:architectures}b respectively. The feedforward inference network
consists of a single proposal layer, $f_{a_t,i_t}^{prop}$. This embedding layer
takes all observe embeddings as input and computes $\eta_t$ as output. Since no
part of its input is dependent on the previously sampled values, unlike the LSTM
architectures, this architecture is completely unable to learn proposal
distributions with dependencies between latent variables before the addition of
an attention mechanism.

To attend to previously sampled variables we add dot-product attention,
described in section~\ref{sec:dpa}, to both the LSTM and feedforward
architectures, as shown in Figure~\ref{fig:architectures}. During training
we build a data structure, $d_\text{k,v,q}$, with associative mappings linking
address/instance pairs $(a,i)$ to key, value and query embedders. The embedders
in $d_\text{k,v,q}$ are constructed dynamically for each new address and
instance pair $(a_t,i_t)$ encountered.

During inference, the queries, keys, and values input to the attention mechanism
at each \texttt{sample} statement are calculated as follows: for the first
\texttt{sample} statement, identified by $(a_1,i_1)$, no previously sampled
variables exist and so the attention module outputs a vector of zeros. Using the
associated key and value embedders in $d_\text{k,v,q}$, the variable sampled,
$x_1$, is embedded to yield a key and a value, $k_1$ and $v_1$. $(k_1,v_1)$ are
kept in memory throughout the trace, allowing fast access for subsequent
\texttt{sample} statements. The second \texttt{sample} statement can attend to
the first sampled variable via $(k_1,v_1)$ using a query. The embedder used for
finding the query takes as input the observe embedding, $f^{obs}(\vy)$, and is
specific to the current address and instance $(a_2,i_2)$. As with the key/value
embedders, the query embedder is found in $d_\text{k,v,q}$. The output of the
attention module is then fed to the LSTM or proposal layer (see
Figure~\ref{fig:architectures}). As for $x_1$, $x_2$ is sampled and embedded
using the embedders stored in $d_\text{k,v,q}$, yielding the key, value pair
$(k_2, v_2)$. This procedure is repeated until the end of the trace, as defined
by the probabilistic program. In the context of higher-order programs, an
address and instance pair may be encountered during inference that has not been
seen during training. In this case the proposal layers are not trained, and so
the standard IC approach is to use the prior as a proposal distribution. For the
same reason, the key/value embedders do not exist and so no keys or values are
created for this $(a_t,i_t)$. This prevents later \texttt{sample} statements
from attending to the variable sampled at $(a_t,i_t)$.

\section{EXPERIMENTS AND RESULTS}
\label{sec:experiments}
We implemented our attention mechanism in, and performed experiments with,
\textit{pyprob} \citep{le2016inference, pyprob}, a PPL designed for IC. For both
experiments the inference networks were trained using Adam~\citep{kingma2014adam}
with hyperparameters $\beta_1=0.9$ and $\beta_2=0.999$ to optimize the loss
given in Eq.~\eqref{eq:loss}. The attention modules we consider use $q = 4$, $k
= 16$ and $v = 8$. We consider feedforward and LSTM architectures both with and
without attention, denoted as $M_{\text{FF w/o ATT}}$, $M_{\text{FF w/ ATT}}$,
$M_{\text{LSTM w/o ATT}}$ and $M_{\text{LSTM w/ ATT}}$.

We test the architectures in two experiments: an illustrative example involving
the estimation of a highly correlated posterior distribution over two latent
variables, and an application of amortized inference to the problem of real-time
circuit fault diagnosis.

We use the effective sample size~\citep{kong1992note} (ESS) of each estimated
posterior as a quantitative metric to describe the quality of the proposal
distribution, and resulting posterior estimate. The effective sample size is not
a perfect metric for the quality of a posterior as, for example, it depends only
on the importance weights of the posterior samples, and so can be high even if
the samples lack diversity and/or miss modes. However, we use it alongside
qualitative plots which show that the improvements in ESS come about through
qualitatively better proposals and estimates of the posterior distributions.

\begin{figure}[t]
  \vskip 0.1in
  \centering
  \includegraphics[width=0.35\textwidth]{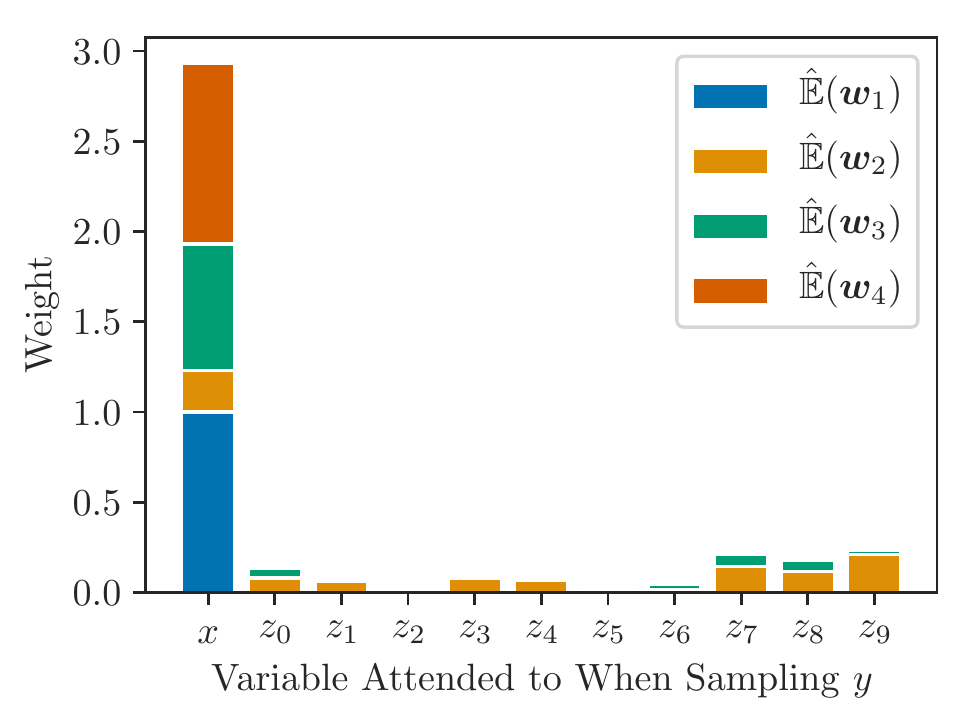}
  \caption{Attention weights used on each previously sampled variable.}
  \label{fig:attention-weights}
\end{figure}

\begin{figure*}[t!]
  \centering
  \includegraphics[width=0.7\textwidth]{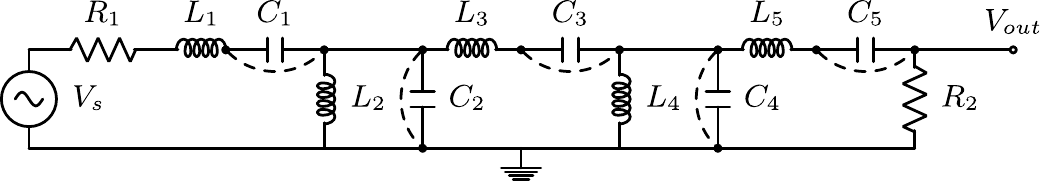}
  \caption{Fifth-order band-pass Butterworth filter with resistors, capacitors
    and inductors denoted by $R,~C$, and $L$ respectively. The dashed lines
    represent possible short circuits. The existence of these short circuits and
    whether or not each component is faulty (represented by a noisy component
    value) or disconnected is sampled according to the generative model. Given
    observations of $V_{\text{out}}$ for various input frequencies, the task is
    to infer a distribution over possible faults such as short circuits and
    poorly connected or incorrectly valued components.}
  \label{fig:butter_filter}
  \vskip -0.2in
\end{figure*}

\subsection{Magnitude of Random Vector}
We first demonstrate the efficacy of our approach on a simple, pedagogical
example. The task is to infer a distribution over two latent variables, $x$ and
$y$, conditioned on $\hat{r}^2$, a noisy observation of $x^2 + y^2$. The
generative model first samples $(x,y)$ coordinates from identical and independent
Gaussian priors: $x\sim\mathcal{N}(0,\sigma_p),~y\sim\mathcal{N}(0,\sigma_p)$. A
noisy estimate of the radius, $\hat{r}$, is then observed with a likelihood
given by $p(\hat{r}^2|x,y)=\mathcal{N}(\hat{r}^2|x^2+y^2,\sigma_l)$. We use
$\sigma_p = 10$ and $\sigma_l = 0.5$, which leads to a tightly peaked posterior
exhibiting circular symmetry, and so a strong dependence between $x$ and $y$.

In the model as just described, where $x$ and $y$ are sampled consecutively, it
is trivial for either the LSTM or the attentive architecture to learn this
relationship. However, we are interested in testing the learning of long-range
dependencies. We therefore consider variations of this model where, after
sampling $x$, and before sampling $y$, some number of ``nuisance'' random
variables are sampled. These nuisance random variables are not used elsewhere in
the program, and so serve only to increase the ``distance'' between $x$ and $y$.
In particular, we consider two programs: one containing 10 ``nuisance''
variables, and one containing 20. Program~\ref{mag2} provides pseudocode for
these programs. All inference networks are trained using $6\times10^5$ traces
with minibatches of size 128. The learning rate is decreased every $2\times10^5$
traces, iterating through $\{ 10^{-3}, 10^{-4}, 10^{-5}\}$.

Figure~\ref{fig:circ_dist} shows $2000$ samples from the proposal distributions
for each program,
$q(x,y|\hat{r}^2)=\int_{z_{1:i}}q(x,y,z_{1:i}|\hat{r}^2)\mathrm{d}z_{1:i}$ for
$i=\{10,20\}$, produced by each architecture. We
see that $M_\text{FF w/o ATT}$ is unable to make proposals capturing the
dependency between $x$ and $y$. This illustrates that allowing the neural
network to ``remember'' (via attention or the LSTM core) the sampled value of
$x$ is necessary to accurately approximate the posterior. We observe that with
only 10 nuisance random variables there is no advantage in using the attention
mechanism compared to an LSTM core. However, when the number of nuisance random
variables increases to 20 the LSTM core is no longer able to capture the
dependency between $x$ and $y$. In contrast, the architectures with attention
mechanisms are unaffected.

\begin{lstlisting}[mathescape=true, caption=Generative model for the magnitude
of a random vector with $M$ nuisance random variables., label=mag2]
def magnitude($\mathrm{obs}$, $M$):
    $x$ = sample(Normal(0, 10))
    for _ in range($M$):
        # nuisance variables to extend trace
        _ = sample(Normal(0,10))
    $y$ = sample(Normal(0,10))
    observe($\mathrm{obs}^2$,
           Likelihood=Normal($x^2 + y^2$, 0.1))
    return $x$, $y$
\end{lstlisting}

Figure~\ref{fig:attention-weights} shows the average attention weights given to
each previously sampled variable (by each of the four queries) when creating a
proposal distribution for $y$. It can be seen that queries 1 and 4 attend solely
to $x$, explaining how the attention mechanism enables the inference network to
capture the long-term dependency, and ignore the nuisance variables.

\subsection{Electronic Circuit Fault Diagnosis}\label{sec:circuits}

For our second experiment, we consider performing inference in a probabilistic
program that imports and uses a pre-existing electronic circuit simulator~\citep{ahkab}.
Specifically, we will consider a Butterworth filter as shown in
Figure~\ref{fig:butter_filter}. This is operated with an input voltage composed
of a $5\si{\volt}$ DC signal and a $1\si{\volt}$ AC signal. The observable output voltages
at different AC frequencies are shown in Figure 4 in the Appendix. The
task is to infer whether or not each component in the Butterworth filter is
faulty given the observed complex-valued output voltage $V_{\text{out}}$ (i.e.
voltage magnitude and phase) at 40 different frequencies. To perform inference
we write a probabilistic program that iterates through each component of the
circuit and samples in the following order: first, whether or not it is
correctly connected to the rest of the circuit. Second, the component value is
sampled from a mixture of a broad uniform distribution and a tightly peaked
Gaussian, both centered on the nominal value. The value is sampled from the
tightly peaked Gaussian with 98\% probability and from the uniform distribution
with 2\% probability. Conceptually, one can interpret the tightly peaked
Gaussian as the distribution given that the component has been correctly made. The broad
uniform distribution represents the distribution for components that are faulty.

To test each inference network, we generate 100 different observations by
running the probabilistic program, and attempt to infer the posterior using each
different network architecture. For each inference network architecture, we
estimate the posterior distribution 5 times using importance sampling with 20
traces each time. Across the 5 estimates, we compute the average ESS, and
average these over all 100 observations. The averaged results were 1.40 for
$M_\text{FF w/o ATT}$, 7.26 for $M_\text{LSTM w/o ATT}$, 8.46 for $M_\text{FF w/
  ATT}$ and 8.35 for $M_\text{LSTM w/ ATT}$.

\begin{figure*}[t!]
  \centering
  \includegraphics[width=\textwidth]{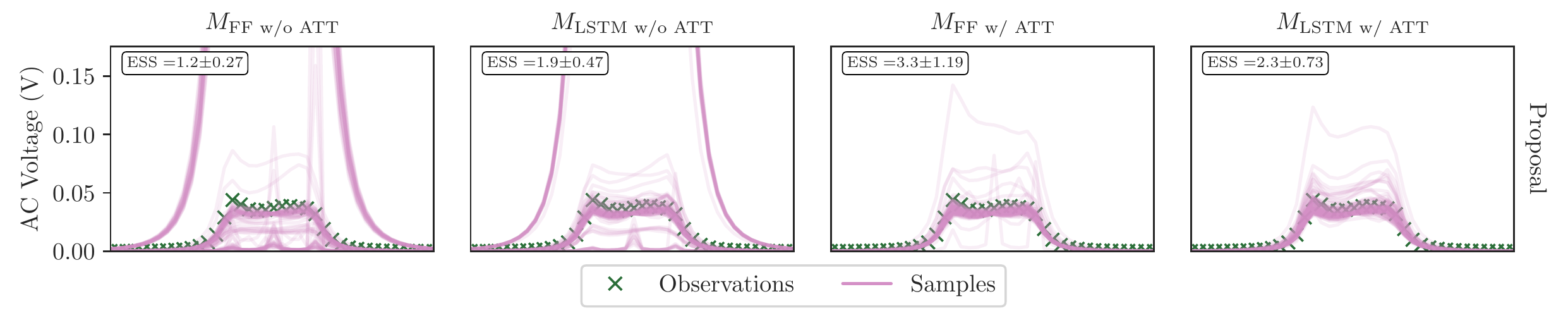}
  \caption{Reconstruction of the output voltage using samples from each proposal
    distribution. In the architectures with attention, the sampled voltages are
    almost all close to the observations (green `x's) whereas, without
    attention, the proposals place high probability in regions which do not fit
    the observations. These better proposal distributions lead the higher
    effective sample sizes shown in each figure (mean and standard deviation,
    calculated with 5 estimates). The proposal distribution is shown using 100
    samples form each architecture. }
  \label{fig:circuit_inference}
\end{figure*}

\begin{figure*}[t!]
  \centering
  \vskip 0.1in
  \includegraphics[width=0.9\textwidth]{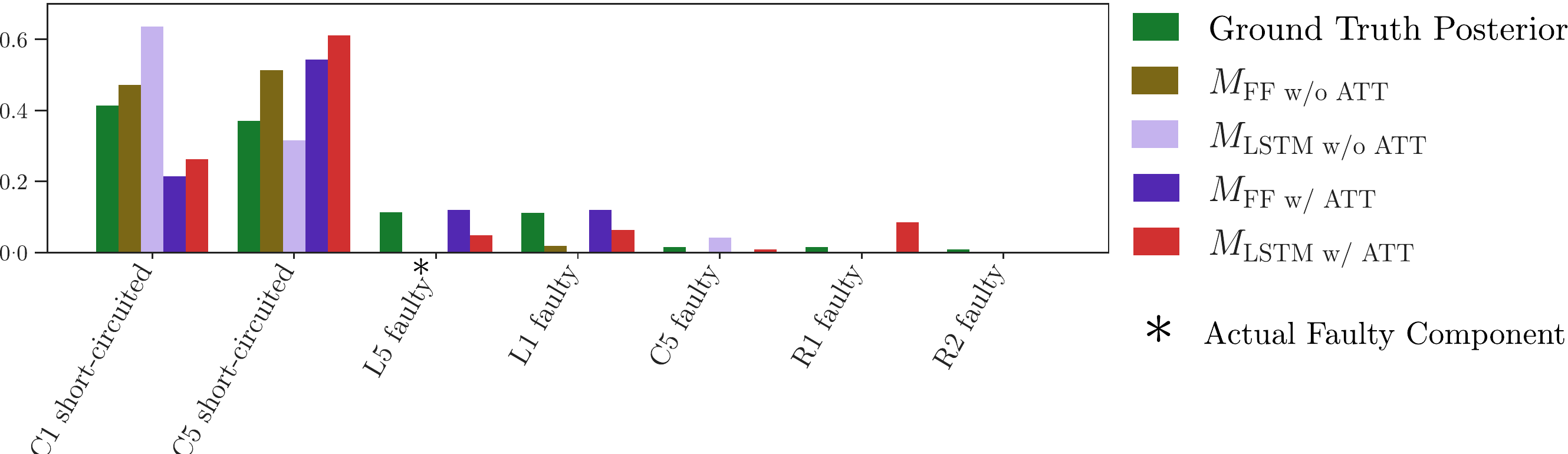}
  \caption{Posterior probability assigned to possible causes of failure. The
    observations used are the same as those in Figure
    \ref{fig:circuit_inference}. In the simulation which generated these
    observations, $L_1$ was the component at fault. However, the observations do
    not contain enough information to infer this exactly, and so being close to
    the ground truth posterior (as shown in green) is the best the inference
    networks can do. The ground truth posterior was inferred using importance sampling with no
    inference network and 17 million samples. The inferred posteriors shown
    were each estimated using $1000$ samples. It can be seen
    that the architectures without attention place very little probability on L1
    or L5 being broken, despite these having a combined probability of about
    20\% according to the true posterior. In particular, they place almost no
    probability on L5 being faulty, even though it was the faulty component in
    the simulation used to create the observed voltages. }
  \label{fig:faulty}
  \vskip -0.2in
\end{figure*}

The attention-based architecture has an 16.5\% higher average ESS than the LSTM
core, showing that the use of attention leads to quantitatively better proposal
distributions. We further find that whenever the observed signal appears to
originate from a correctly working Butterworth filter, all architectures seem to
produce reasonable predictive posterior distributions - i.e. the distribution of
the voltage signal generated by the sampled latent variables. However, the
attention-based architectures yield a higher average ESS with only a few
exceptions. When the observed signal clearly originates from an erroneous
filter, $M_{\text{FF w/o ATT}}$ produces predictive posterior distributions
which poorly fit the observed data. The LSTM-based architecture produces better
predictive posterior distributions but these are still significantly worse than
the distributions produced by the attention-based architecture in almost all
cases where the filter is broken. Figure~\ref{fig:circuit_inference} shows
inference performance for one such observation originating from a filter in
which the component is faulty. We plot voltages generated according to the
sampled latent variables from the predictive proposal distributions using each
architecture. The proposals from $M_{\text{FF w/ ATT}}$ and $M_{\text{LSTM w/
    ATT}}$ are clustered near to the observations, whereas the proposals from
$M_{\text{FF w/o ATT}}$ and $M_{\text{LSTM w/o ATT}}$ produce many proposals
that do not fit the observations.

We suspect that these outliers occur due to the inability of $M_{\text{FF w/o
    ATT}}$ and $M_{\text{LSTM w/o ATT}}$ to learn long-range dependencies. For
example, an output voltage of zero could be explained by a number of different
faults (e.g. a short-circuit across $C_2$ or across $C_4$). If the resulting
dependency between these can be learned, the proposals could consistently
predict that only one is broken (predicting more would be unlikely due to the
strong prior on parts working). However, if the dependency is not captured, the
proposals would be prone to predicting that zero or multiple components are
broken. This interpretation is supported by Figure \ref{fig:circuit_inference},
where both architectures without attention are seen to sometimes propose an
output voltage corresponding closely to a working
circuit. 

Figure~\ref{fig:faulty} shows an example of the posterior distributions inferred
over possible faults by each architecture. For this purpose, a component is
considered faulty when its value is outside of a 0.3\% tolerance of its nominal
value. L5 was at fault when the observations used were generated, but the
observations do not allow this to be inferred with confidence as other failures
could have produced the same observations. This is reflected in the ground truth
posterior placing little probability on L5 being faulty. The architectures with
attention manage to most closely fit the ground truth posterior. In particular,
they assign probability to L5 being faulty whereas the other architectures
assign very little.

\section{DISCUSSION AND CONCLUSION}
\label{sec:conclusion}
We have demonstrated that the standard LSTM core used in IC can fail to capture
long-range dependencies between latent variables. To address this, we have
proposed an attention mechanism which enables the inference network to attend to
the most salient previously sampled variables in an execution trace. We further
demonstrate that attention improves inference in a practical application of IC,
yielding better proposal distributions and thus posteriors with an effective
sample size 16.5\% higher than when using the LSTM core.

Our work leads to several avenues of future research: first, while we
demonstrate the efficacy of an architecture using only an attention mechanism
relative to an architecture using only an LSTM, further investigation is needed
to determine whether there could be additional benefits from using both in
conjunction. Additionally, this work raises the possibility of extending the
usage of such attention mechanisms to attend to the observations in a generic
way. This could enable the inference network to filter out noise in the
observations and reduce computation time. Furthermore, the inference compilation
framework currently requires a fixed number of observations. An attention
mechanism over the observations may allow this requirement to be relaxed.

\subsubsection*{Acknowledgements}
We acknowledge the support of the Natural Sciences and Engineering Research
Council of Canada (NSERC), the Canada CIFAR AI Chairs Program, Compute Canada,
Intel, and DARPA under its D3M and LWLL programs.

\bibliographystyle{apalike}
\bibliography{bibfile.bib}

\end{document}